\pdfoutput=1

\documentclass[11pt]{article}

\usepackage{acl}
\usepackage{hyperref}

\usepackage{times}
\usepackage{latexsym}
\usepackage{amsmath}
\usepackage{subfigure}

\usepackage{graphicx}
\graphicspath{{./fig}}
\usepackage[T1]{fontenc}

\usepackage[utf8]{inputenc}

\usepackage{microtype}

\usepackage{xcolor}
\usepackage{multirow}
\usepackage{booktabs}   
\usepackage{nicematrix}

\newif\ifcomments

\commentstrue

\ifcomments
\newcommand{\comments}[1]{#1}
\else
\newcommand{\comments}[1]{}
\fi

\newcommand{\papertitle}{Semi-supervised Neural Machine Translation with Consistency Regularization for Low-Resource Languages}
\title{\papertitle}

\author{Viet H. Pham$^{\textsuperscript{1}}$, Thang M. Pham$^{\textsuperscript{2*}}$ , Giang Nguyen$^{\textsuperscript{2*}}$, Long Nguyen$^{\textsuperscript{1}}$, Dien Dinh$^{\textsuperscript{1}}$ \\
$^{\textsuperscript{1}}$University of Science, VNUHCM\\$^{\textsuperscript{2}}$Auburn University\\
\texttt{\{viethungpham0304, giangnguyenbkhn, long.hb.nguyen\}@gmail.com}\\
\texttt{thangpham@auburn.edu}\\
\texttt{ddien@fit.hcmus.edu.vn}
}

\begin{document}

\maketitle

\renewcommand{\thefootnote}{\fnsymbol{footnote}}
\footnotetext[1]{Equal contributors}
\renewcommand{\thefootnote}{\arabic{footnote}}
\begin{abstract}
The advent of deep learning has led to a significant gain in machine translation. However, most of the studies required a large parallel dataset which is scarce and expensive to construct and even unavailable for some languages. This paper presents a simple yet effective method to tackle this problem for low-resource languages by augmenting high-quality sentence pairs and training NMT models in a semi-supervised manner.
Specifically, our approach combines the cross-entropy loss for supervised learning with KL Divergence for unsupervised fashion given pseudo and augmented target sentences derived from the model.
We also introduce a SentenceBERT-based filter to enhance the quality of augmenting data by retaining semantically similar sentence pairs.
Experimental results show that our approach significantly improves NMT baselines, especially on low-resource datasets with 0.46--2.03 BLEU scores.
We also demonstrate that using unsupervised training for augmented data is more efficient than reusing the ground-truth target sentences for supervised learning \cite{BT}.
\end{abstract}

\section{Introduction}
Neural Machine Translation (NMT) models are based on the encoder-decoder architecture \cite{sutskever, gehring, vaswani, bahdanau, Luong2015EffectiveAT}, have been widely used to improve translation quality effectively. 

However, training a neural machine translation (NMT) model requires abundant parallel data and computational costs. Large parallel corpora are only available for some of the high-resource languages such as English, France, German, and Romanian. 
Regarding low-resource languages, existing corpora with high quality and quantity are unavailable. 
Hence, for NMT models to work well with low-resource languages, multiple studies are trying to take advantage of both data synthesis and model customization approaches.

Data synthesis with the domination of back-translation (BT) approaches has been widely used since it was first proposed for NMT in 2016 \cite{sennrich-etal-2016-improving}. Since then, numerous studies have focused on improving NMT models with BT, as the major data augmentation method. However, the NMT models do not always work well with input perturbations induced by data augmentation. It may lead to a situation in which the NMT models dramatically decrease translation quality. Therefore, we suggest training NMT models that are invariant to data augmentations.

\begin{figure*}
    \centering
    \includegraphics[scale=0.7]{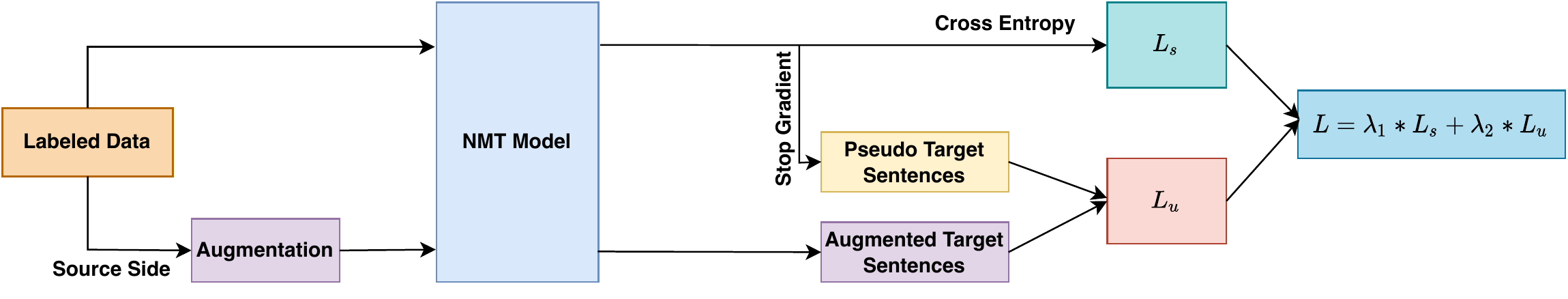}
    \caption{Workflow of our proposed method. Labeled data are fed into NMT model for computing supervised loss (i.e., cross-entropy loss and soft pseudo-labels -- the upper branch). On the other hand, the source side of labeled data is augmented and fed into the same model to get ``Augmented Target Sentences''. Then, we compute the closeness of ``Augmented Target Sentences'' and ``Pseudo Target Sentences'' of the same sentence (e.g. like pseudo-labeling in \cite{nguyen2019contcap, nguyen2021explaining}). Consequently, the final objective is the summation of two losses.}
    \label{fig:overview}
\end{figure*}

This paper proposes a semi-supervised approach to tackle the data-shortage problem of NMT models.
Our approach jointly trains NMT model using parallel and monolingual data. The main idea is to use one unsupervised objective for the monolingual data beyond traditional supervised loss.
We also leverage SentenceBERT - a multilingual semantic model to filter out unqualified augmented sentences that potentially hurt the model performance.
The overview of our method is shown in Figure \ref{fig:overview}.
Our main contributions can be summarized as follows: 
\begin{itemize}
    \item We propose a semi-supervised learning method to use augmented data to improve the translation quality compared to strong baselines.
    \item We introduce a semantic filter to enhance the quality of generated data derived from augmentation methods.
\end{itemize}

\section{Related Work}

Recently, several methods exploiting monolingual data to improve the performance of the NMT models. \citealp{cheng-etal-2016-semi} proposed a semi-supervised approach that reconstructs the monolingual corpora using an encoder-decoder model. Methods by \citealp{skorokhodov-etal-2018-semi} used transfer learning to utilize the knowledge from language models for translation systems. With the use of supervised source domain data, \citealp[]{Jin2020UnsupervisedDA} developed domain adaptation scenarios for NMT to enhance the performance of a translation model on a destination domain without parallel data.

Back-translation (BT) \cite{BT, Sennrich2016ImprovingNM}, an alternative to leverage monolingual data, trains a target-to-source system to generate additional synthetic parallel data from the monolingual data.

The closest method to our work is FixMatch \cite{Sohn2020FixMatchSS} which adopted Consistency Regularization \cite{cr} to semi-supervised learning and achieved state-of-the-art results on image classification tasks. 
Our method differs in how we adapt this concept for translation tasks: KL Divergence to compute the semantic distance between two output distributions (Figure~\ref{fig:overview}).
\section{Machine Translation with Consistency Regularization}
As shown in Figure \ref{fig:overview}, the proposed method is very simple. For each source-target pair, we applied the back-translation method on the source sentence to obtain its augmented version and used to compute the KL divergence between the Pseudo Target and Augmented Target sentence. The final loss function is the combination of traditional cross-entropy loss and the KL divergence as follow:

\begin{equation}
    L = \lambda_1L_{CE} + \lambda_2L_{KL}
\end{equation}
in which $\lambda_1$ and $\lambda_2$ is the hyperparameters that balance the supervised and unsupervised loss.

\section{Experiments}
\subsection{Datasets}
\begin{table}[]
    \centering
    \begin{tabular}{|l|c|c|c|}
    \hline
        \textbf{Dataset} & \textbf{Train} & \textbf{Dev} & \textbf{Test} \\\hline
        WMT14 En-De &4.56M&3K&2.6K \\\hline
        IWSLT15 &&&\\
\hspace{5mm}+ En-Vi &133K &2K &2K \\\hline
        IWSLT17&&&\\
\hspace{5mm}+ En-Fr &236K& 888 &1K \\
\hspace{5mm}+ En-De &209K&888 &1K \\\hline
    \end{tabular}
    \caption{The statistics of datasets. En-De: English to German; En-Vi: IWST15 English to Vietnamese; En-Fr: IWSLT17 English to France; En-De: IWSLT17 English to German. Unit: number of sentences.}
    \label{tab:data_stat}
\end{table}
We evaluated our method in two settings: large-scaled corpus with  WMT14 English-German \footnote{https://www.statmt.org/wmt16/translation-task.html}  and low-scaled corpus with IWSLT dataset (IWSLT17 English-France, English-German and IWSLT15 English-Vietnamese). All statistical information is shown in Table \ref{tab:data_stat}.

\textbf{Unlabeled data construction}: We applied back-translation on the English side for each language pair to generate augmented version of input sentences. Then, all duplicate examples in the original and augmented data were filtered out and only retained those such that:
\begin{equation}
    \frac{\text{len}(x_i)}{\text{len}(u_i)} \in [\alpha_1, \alpha_2]
\end{equation}
This criteria makes sure that the length\footnote{Based-on words} of two sentence is not much different. In our experiments, we set $\alpha_1 = 0.7$ and $\alpha_2 = 1.4$.
Then, we use sbert\footnote{https://www.sbert.net/} to eliminate those sentence pairs that has a low similarity score ( $< 0.5$ in our experiments). Because we used the English side of each dataset to augment, $\mu$ is set to $1$.

After constructing a quality unlabeled dataset, Byte-Pair Encoding (BPE) \cite{bpe} was applied with 3000 operations for the IWSLT dataset, 30K for WMT14 English-German.

\subsection{Models and Hyperparameters}
As our method does not change model architectures, we can use any NMT models like Transformer, BART, LSTM, etc. as backbones. In our experiments, we used Transformer-based and BART-based due to hardware limitation. Hyperparameters are described in Appendix \ref{training_details}.

\begin{figure*}
    \centering
    \includegraphics[width=\linewidth]{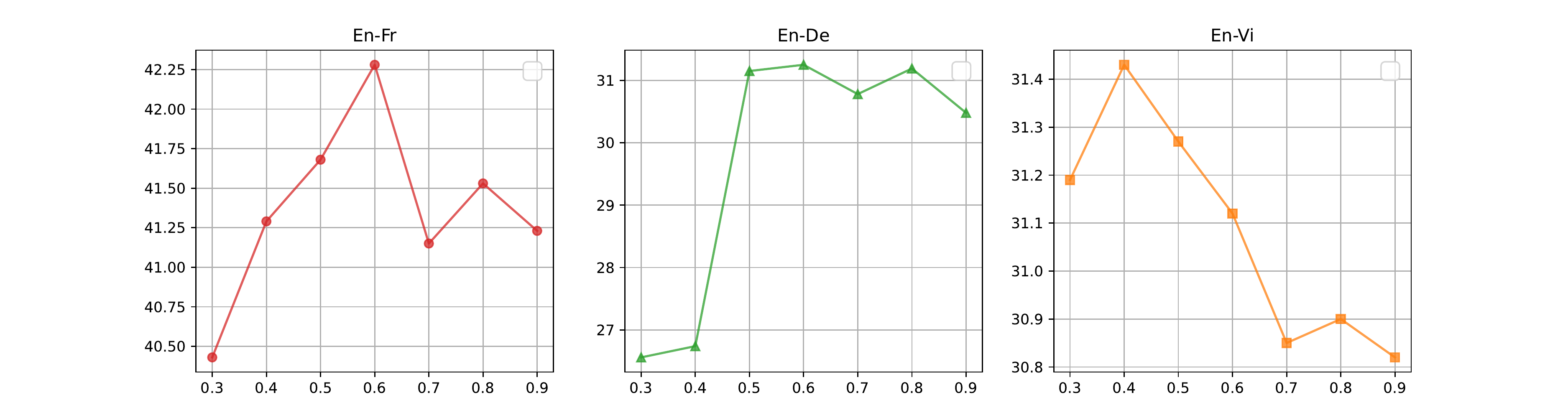}
    \caption{Model's performance on three IWSLT dataset when varying $\lambda_1$. The vertical axis presents BLEU4 score, horizontal axis presents the value of $\lambda_1$, which is the weight of the cross entropy. In this experiment, we still kept constraining $\lambda_1 + \lambda_2 = 1.0$.}
    \label{fig:loss_weight}
\end{figure*}
\subsection{Results}

We compare our proposed method with various strong baselines such as Transformer \cite{NIPS2017_3f5ee243}, BART \cite{Lewis2020BARTDS}, Transformer Cycle \cite{Takase2021LessonsOP} and back-translation \cite{BT}. Table \ref{result} shows BLEU score of each method on test set. As can see from the table, our method significantly improved the translation quality from +0.4 (IWSLT17 En-De) to +2.0 (IWSLT15 En-Vi) BLEU score for all dataset. The most improvement is En-Vi which is $2.03$ BLEU higher than the vanilla Transformer. Interestingly, our method has also outperformed the back-translation which is known as one of the best method for utilizing unlabeled dataset in machine translation, pushing at least $+1.88$ BLEU. For other low-resource dataset such as En-Fr and En-De, our model has consistently achieved better BLEU score which is higher $1.18$ and $0.46$ compared to Transformer baseline, respectively.

On the larger dataset, WMT14 En-De, the highest result is marked in bold, this proves that the Consistency Regularization framework works well on various scales of datasets (i.e. small and large).

\begin{table}
\centering
\resizebox{0.485\textwidth}{!}{
\begin{NiceTabular}{|l|c|c|c|c|}
\hline
\multirow{2}{*}{Model} & \multicolumn{2}{c}{IWSLT17} & IWSLT15 & WMT14 \\
\cmidrule{2-3}
& \textbf{En-De} & \textbf{En-Fr} & \textbf{En-Vi} & \textbf{En-De}\\
\hline
Transformer-based & 30.79 & 41.10 & 29.40 & 28.98 \\\hline
BART-based & 28.81 & 40.37 & 29.65 & 28.77 \\\hline
Transformer Cycle & & & &\\
\hspace{5mm}+ Cycle &29.76  &40.87 &30.47 &28.32 \\
\hspace{5mm}+ Sequence &29.84  &40.80  &30.72 &29.01 \\
\hspace{5mm}+ Cycle reverse &29.29 &40.52 &30.07 &28.24 \\\hline
Back-translation & 30.79 & 40.49 & 29.55 & 29.30 \\\hline
\textbf{Ours} & &&& \\
\hspace{5mm}+ Transformer & \textbf{31.25} & \textbf{42.28}  & \textbf{31.43} & \textbf{29.39} \\
\hspace{5mm}+ BART-based & 30.39 & 41.30  & 29.96 & 25.90 \\\hline

\end{NiceTabular}
}
\caption{\label{result} BLEU scores on each dataset. We report our results with various backbones. We generate translations with a beam size of 5 and length penalty of 0.6.
}
\end{table}

\subsection{Ablation Study}
Since our proposed method contains two terms of loss, it is important to know what is the optimal values of $\lambda_1$ and $\lambda_2$
. Moreover, we found that the augmentation strategy also contributed significantly to the performance gain. In this section, we conduct experiments to answer two questions:
\begin{itemize}
    \item How importance does the loss weight contribute to final results?
    \item How does augmentation strategies affect the translation quality?
\end{itemize}
We used three IWSLT dataset (En-Vi, En-De and En-Fr) for all ablation studies in this section.
\subsubsection{The Loss Weight}
We study the interaction between two components of our loss function: supervised and unsupervised by adjusting their weight and still constrained $\lambda_1 + \lambda_2=1.0$. As shown in Figure~\ref{fig:loss_weight}, our models on three IWSLT dataset reached peak performance at $\lambda_1=\{0.4, 0.5, 0.6\}$, which is similar to the proportion of the labeled and unlabeled examples.
Moreover, the BLEU score is prone to decrease significantly when $\lambda_1 >0.6$, contrasting to the dramatically increase when $\lambda_1 \le 0.6$. This shows that the ratio between $\lambda_1$ and $\lambda_2$ should be similar to the proportion of labeled and unlabeled samples.

\subsubsection{Augmentation Strategies}

\begin{table}[hbt!]
    \centering
    \begin{NiceTabular}{|l|c|c|c|}
    \hline
        & \multicolumn{2}{c}{IWSLT17} & IWSLT15 \\
        \cmidrule{2-3}
         & En-De & En-Fr & En-Vi \\\hline
        BackTranslation &\textbf{31.25}&\textbf{42.28}&\textbf{31.43} \\\hline
        WordDropout &30.68&41.27& 26.93 \\\hline
        Synonym&30.13&40.90& 28.11\\\hline
        Word order &30.25&41.14&28.06  \\\hline
        \end{NiceTabular}
    \caption{Experiments on different augmentation methods. For WordDropout, Synonym and WordOrder, we randomly drop $p=0.2$ tokens in the sentence, used pretrained RoBERTa to substitute synonyms and randomly choose two adjacent words in the sentence and swap their positions, respectively.}
    \label{tab:aug_strategy}
\end{table}
As presented in Table \ref{tab:aug_strategy}, we found that model performance is relatively sensitive to the augmentation methods.
Specifically, WordDropout \cite{iyyer-etal-2015-deep} and WordOrder have lower BLEU scores compared to Back-translation and Synonym. One possible explanation is that the breaking of syntactic information leading to the change of meaning which results in the KL Divergence between two sentences being large and translating incorrectly. Meanwhile, Back-Translation and synonym only replace words have the same meaning, resulting in both syntactic and semantic aspect of the sentence still remain unchanged. 

\section{Conclusion}
We presented a simple semi-supervised approach for neural machine translation which utilizes both parallel data and monolingual data. Our method has consistently outperformed baselines on both low scale and large scale dataset. We also find that the choice of augmentation strategies is important which contribute directly to BLEU score. Future work will improve our
approach in terms of model architecture and augmentation methods as well as explore the effectiveness in different settings for low-resource corpora.



\section*{Acknowledgements}
We would like to thank Viet Phan, an AI Engineer at VinAI Research for your valuable discussion and comments.

\bibliography{anthology,custom}
\bibliographystyle{acl_natbib}

\newcommand{\beginsupplementary}{%
    \setcounter{table}{0}
    \renewcommand{\thetable}{A\arabic{table}}%
    \setcounter{figure}{0}
    \renewcommand{\thefigure}{A\arabic{figure}}%
}

\beginsupplementary%
\appendix


\newcommand{\toptitlebar}{
    \hrule height 4pt
    \vskip 0.25in
    \vskip -\parskip%
}
\newcommand{\bottomtitlebar}{
    \vskip 0.29in
    \vskip -\parskip%
    \hrule height 1pt
    \vskip 0.09in%
}

\newcommand{\suptitle}{Appendix for:\\\papertitle}

\newcommand{\maketitlesupp}{
    \newpage
    \onecolumn
        \null
        \vskip .375in
        \begin{center}
            \toptitlebar
            {\Large \bf \suptitle\par}
            \bottomtitlebar
            \vspace*{24pt}
            {
                \large
                \lineskip=.5em
                \par
            }
            \vskip .5em
            \vspace*{12pt}
        \end{center}
}

\maketitlesupp%

\section{Training Details} \label{training_details}
We implemented the Transformer model in PyTorch using fairseq toolkit\footnote{https://github.com/facebookresearch/fairseq}. For model architecture, we used the same hyperparameters for all experiments, i.e., 4 Transformer blocks, 8 attention heads, word embeddings and feed-forward layers dimensions are 1024 and 2048, respectively. Dropout is set to 0.2, and we averaged the checkpoints of the last ten epochs. Adam optimizer \cite{Kingma2015AdamAM} is adopted with $\alpha = 0.9, \beta = 0.98$ and a leaning rate of 0.0005 and inverse square root scheduler with the initial value of $10^{-7}$. All models used label smoothing uniform prior distribution over the vocabulary $\epsilon = 0.1$. We used a dynamic batch size with total tokens are 3000 per batch. To avoid overfitting, we used weight decay of 0.0001 and trained in 30 epochs for all models.\\
\end{document}